\documentclass{article}
\usepackage{spconf}
\usepackage[cmex10]{amsmath}
\usepackage{amsthm}
\usepackage{amssymb}
\usepackage{mathrsfs}
\usepackage{graphicx}
\usepackage{float}
\usepackage{array}
\usepackage{epstopdf}
\usepackage{multirow}

\newcommand{\argmin}{\arg\!\min}
\newcommand{\argmax}{\arg\!\max}

\title{O\MakeLowercase{n} T\MakeLowercase{he} P\MakeLowercase{ower} \MakeLowercase{of} J\MakeLowercase{oint} W\MakeLowercase{avelet}-DCT F\MakeLowercase{eatures} \MakeLowercase{for} M\MakeLowercase{ultispectral} P\MakeLowercase{almprint} R\MakeLowercase{ecognition}}

\name{Shervin Minaee and AmirAli Abdolrashidi}
\address{Electrical and Computer Engineering Department, New York University, USA.}

\begin{document}
%\ninept
%
\maketitle
\begin{abstract}
Biometric-based identification has drawn a lot of attention in the recent years. Among all biometrics, palmprint is known to possess a rich set of features. In this paper we have proposed to use DCT-based features in parallel with wavelet-based ones for palmprint identification. PCA is applied to the features to reduce their dimensionality and the majority voting algorithm is used to perform classification. The features introduced here result in a near-perfectly accurate identification. This method is tested on a well-known multispectral palmprint database and an accuracy rate of 99.97-100\% is achieved, outperforming all previous methods in similar conditions.
\end{abstract}
%
%\begin{keywords}
%One, two, three, four, five
%\end{keywords}
%
\section{Introduction}
\label{sec:intro}
To personalize an experience or make an application more secure, we may need to be able to distinguish a person from others. To do so, many alternatives are available, such as keys, passwords and cards. The most secure options so far, however, are biometric features. They are divided into behavioral features that the person can uniquely create (signatures or walking rhythm), and physiological characteristics (fingerprints and iris pattern). Many works revolve around identification, verification and categorization of such data including but not limited to fingerprints \cite{Fingerprint}, palmprints \cite{Palm}, faces \cite{Face} and iris patterns \cite{Iris}.

%In many identification systems, there is a database containing desired samples. To identify, the device takes a sample of the entry, extracts its features and compares them to the others. Usually there is a threshold present that separate false data, or non-existence of the person in the database, from true. If the person is present in the source, the program finds the closest match and allows the person to use the protected service. At the same time, the authorization may be recorded for future use, or the new sample might replace one of the old ones for the same process in the future.

Palmprint is among the most popular biometrics due to the many features it possesses and its stability over time. To use palmprints to such end, two widespread methods exist; either transforming the images into another domain like Fourier, DCT, wavelet or Gabor; or attempting to extract the lines and the geometrical characteristics from the palms. Many transform-based approaches exist, such as \cite{Gabor}, in which Zeng utilized two-dimensional Gabor-based features and a nearest-neighbor classifier for palmprint recognition, \cite{wavelet} in which Wu presented a wavelet-based approach for palmprint recognition and used wavelet energy distribution as a discriminant for the recognition process and \cite{PCA+GWR} in which Ekinci proposed a wavelet representation approach followed by kernel PCA for palmprint recognition.
Among notable line-based approaches is \cite{crease} where Cook proposed an automated flexion crease identification algorithm using image seams and KD-tree nearest-neighbor searching which results in a very high recognition accuracy.

There have also been notable developments in the more recent works.
In \cite{hol}, Jia proposed a new descriptor for palmprint recognition called histogram of oriented lines (HOL) which is inspired by the histogram of oriented gradients (HOG) descriptors.
The work presented in \cite{quaternion} by Xu involves a quaternion principal component analysis approach for multispectral palmprint recognition with a high accuracy. In \cite{palm1}, Minaee proposed to use a set of statistical and wavelet features to perform the identification task. 
In \cite{texture}, Minaee proposed a set of textural features derived from the co-occurrence matrices of palmprint blocks and with the use of majority voting, achieved a highly accurate identification.

Most of the palmprint recognition systems consist of four general steps: image acquisition, preprocessing, feature extraction and template matching. These steps are shown in the block diagram in Figure 1.

\begin{figure}[1 h]
\begin{center}
    \includegraphics [scale=0.45] {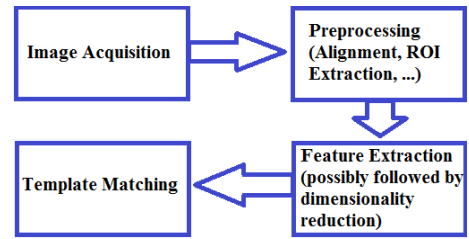}
\end{center}
\vspace{-0.3cm}
  \caption{Block diagram of biometric recognition scheme}
\end{figure}

Images can be acquired by different devices, such as CCD cameras, digital cameras and scanners.
In our work, we have used the multispectral palmprint database which is provided by Polytechnic University of Hong Kong \cite{verification}, \cite{database}.
Four sample palmprints from this dataset are shown in Figure 2.

\begin{figure}[2 h]

\begin{center}
    \includegraphics [scale=0.6] {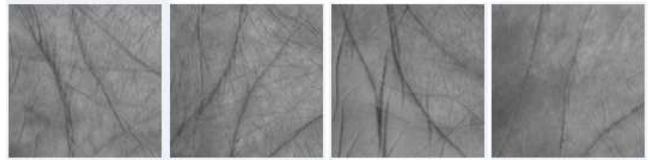}
\end{center}
\vspace{-0.2cm}
  \caption{Four sample palmprint images from PolyU dataset}
\end{figure}

In the feature extraction step, we have used a combined set of DCT and wavelet features. PCA is applied to the features to reduce their dimensionality. In spite of the simplicity of these features, they prove to be quite effective for multispectral palmprint recognition.
After feature extraction, we have used the majority voting scheme and minimum distance classifier to match and identify palmprints.
The proposed algorithm is very fast and can be implemented in electronic devices in conjunction with energy-efficient algorithms
\cite{hoseini1}, \cite{hoseini2}.

The following sections of this paper explain what and how features are used in this classification. Section \ref{SectionII} describes the proposed set of features. Section \ref{SectionIII} contains an explanation of our classification technique. Results of our experiments and comparisons with other works are in Section \ref{SectionIV} and  the conclusion is in Section \ref{SectionV}.

\section{Features}
\label{SectionII}
Feature extraction is a primary step in data analysis, and the information that features provide is correlated with the accuracy of the algorithm. Highly discriminating features usually have a large variance across different classes of target values and a small variance across samples of each class.
There are many approaches used for feature extraction \cite{feature}. One approach is to attempt to automatically derive the useful set of features from a set of training data by projecting it onto discriminative components such as PCA and ICA. The other approach is to use hand-crafted features such as SIFT and HOG (Histogram of Oriented Gradients), or features from transform domain such as wavelet \cite{palm1}. Geometric features are also popular in many medical applications \cite{chro}. Sparse representation has also been used for extracting features in image classification task \cite{hojat1}, \cite{hojat2}. One can also use dictionary learning framework to learn a good set of features from a set of training data \cite{joneidi1}-\cite{joneidi2}.  
%In recent years, features based on deep representation has also been very popular \cite{iris1}.

Here a combined set of DCT- and wavelet-based features is used to perform multispectral palmprint identification. These features are extracted from small patches of each image and subsequently, features of different patches are concatenated to form the final feature matrix of each image. PCA can also be applied to the features for dimensionality reduction.

\subsection{DCT Domain Features}
Discrete cosine transform (DCT) has many applications in various areas of image processing including compression and denoising \cite{dct}. Because of its energy compaction property, most of the image information tends to be concentrated in a few DCT coefficients and makes it favorable for image compression applications.

Suppose we have a 2D discrete function $f(m,n)$ of size $M\times N$. Its 2D DCT is defined as:
\footnotesize
\begin{gather*}
F(u,v)= \alpha_u \alpha_v \sum_{m=0}^{M-1} \sum_{n=0}^{N-1} f(m,n) cos(\frac{\pi(2m+1)u}{2M})cos(\frac{\pi(2n+1)v}{2N})
\end{gather*}
\normalsize
where $0 \leq u <M-1$, $0 \leq v <N-1$ and:
\begin{equation*}
  \alpha_u=\begin{cases}
    \sqrt{1/M} & \text{if $u=0$}\\
    \sqrt{2/M} & \text{otherwise}
  \end{cases}  \ , \ \    \alpha_v=\begin{cases}
    \sqrt{1/N} & \text{if $v=0$}\\
    \sqrt{2/N} & \text{otherwise}
  \end{cases}
\end{equation*}

To extract DCT features from palmprints, every palmprint is divided into non-overlapping blocks of size 16$\times$16 and the 2D DCT of each block is computed. As we know, for most of camera-captured images, the majority of the energy is contained in the upper right subset of DCT coefficients. Because of that, the first 9 DCT coefficients in the zig-zag order are selected as DCT features. These features are shown in the following matrix.

$$
\begin{bmatrix}
f_0 & f_1 & f_5 & f_6 & \cdots \\
f_2 & f_4 & f_7 & ~ & ~\\
f_3 & f_{8} & \ddots & ~ & ~\\
f_{9} & ~ & ~ & ~ & ~ \\
\vdots &~&~&~&~
\end{bmatrix}_{16\times16}
$$
%Note that the zero frequency coefficient ($f_0$) will not be used due to it being the same as the mean value of intensities.

One can also keep more than 9 DCT coefficients or can also make use of all DCT coefficients. However, based on our experiments, using more than 9 coefficients does not provide us with significant improvement.

\subsection{Wavelet Domain Features}
Wavelet is a very popular tool for a variety of signal processing applications such as signal denoising, signal recovery and image compression \cite{mallat}. Perhaps JPEG2000 \cite{jpeg} is one of the most notable examples of wavelet applications.
In our feature extraction procedure, the images are first divided into $16\times 16$ non-overlapping blocks. Then the 2D-wavelet decomposition is performed up to three stages, and in the end, 10 sub-bands are produced. The energy of wavelet coefficients in these subbands are used as the wavelet features (the LL subband of last stage is not used here).
The summary of our wavelet feature extraction stage is presented in the following algorithm:

\begin{enumerate}
\item	Divide each palm image into $16\times 16 $ non-overlapping blocks;

\item	Decompose each block up to 3 levels using Daubechies 2 wavelet transform;

\item	Compute the energy of each subband and treat it as a feature.
\end{enumerate} 

After computation, there will be a total of 18 different features (9 DCT plus 9 wavelet) for each block which can be combined in a vector together: $\textbf{f}=(f_1,f_2,...,f_{18})^\intercal$. It is necessary to find the above features for each image block. If each palm image has a size of $W \times H$, the total number of non-overlapping blocks of size $16\times 16 $ will be:
\begin{gather*}
M=\frac{W \times H}{256}
\end{gather*}
Therefore there are $M$ such feature vectors, $\textbf{f}^{(m)}$. Similarly, they can be put in the columns of a matrix to produce the feature matrix of that palmprint, $\textbf{F}$:
\begin{gather*}
\textbf{F}=[\textbf{f}^{(1)} ~  \textbf{f}^{(2)} \cdots  ~\textbf{f}^{(M)}]
\end{gather*}
There are a total of 1152 features for each palmprint image. Using all of the 1152 features may not be efficient for some applications. In those cases, dimensionality reduction techniques can be used to reduce the complexity.

\subsection{Principal Component Analysis}
Principal component analysis (PCA) is a powerful algorithm used for dimensionality reduction \cite{PCA}. Given a set of correlated variables, PCA transforms them into another domain such the transformed variables are linearly uncorrelated. This set of linearly uncorrelated variables are called principal components. PCA is usually defined in a way that the first principal component has the largest possible variance, the second one has the second largest variance and so on. Therefore after applying PCA, we can keep a subset of principal components with the largest variance to reduce the dimensionality.
PCA has a lot of applications in computer vision and neuroscience. Eigenface is one representative application of PCA in computer vision, where PCA is used for face recognition.

Without going into too much detail, let us assume we have a dataset of $N$ palmprint images and $\{f_1,f_2,...,f_N\}$ denote their features. Also let us assume that each feature has dimensionality of $d$. To apply PCA, we first need to remove the mean value of the features as  $z_i= f_i- \bar{f}$ where $\bar{f}= \frac{1}{N} \sum_{i=1}^m f_i$.
Then the covariance matrix of the centered images is calculated:
\begin{gather*}
C= \sum_{i=1}^m z_i z_i^T 
\end{gather*}
Next the eigenvalues $\lambda_k$ and eigenvectors $\nu_k$ of the covariance matrix $C$ are computed. Suppose $\lambda_k$'s are ordered based on their values. Then each $z_i$ can be written as $z_i= \sum_{i=1}^d \alpha_i \nu_i$.
By keeping the first $k (\ll d)$ terms in this summation, we can reduce the dimensionality of the data by a factor of $\frac{k}{d}$ and derive new feature representation as $\hat{z_i} $.
By keeping $k$ principal components, the percentage of retained energy will be equal to $\frac{\sum_{i=1}^k \lambda_i}{\sum_{i=1}^d \lambda_i }$.

\section{Majority Voting Classifier}
\label{SectionIII}
After the features are extracted, a classifier is required to match the most similar image in the data set to the test subject. There are different classification algorithms that can be used. Some of the most widely used include minimum-distance classifier, neural networks and support vector machines. These algorithms usually have some parameters which need to be tuned. The parameter tuning is usually done by minimizing a cost function on the training set. If the dataset is large enough, the cost function is basically the training error. However if the data set is small, the cost function should have two terms: one term tries to minimize the error; and the other term tries to minimize the risk of over-fitting. One such a work is studied in \cite{mtbi}.

Here we have used the majority voting algorithm. It is performed by individual predictions by every feature followed by adding all the votes to determine the outcome.
One can also use weighted majority voting where each feature is given a weight in the voting process.
The weight of each feature is usually related to the single feature accuracy in the classification task; the more it can successfully predict on its own, the greater weight it is given. Here we have assigned similar weights to all features to make the algorithm parameters independent of the dataset.

In our classifier, first the training images' features are extracted. Then, the features of the test sample are extracted and the algorithm searches for a training image which has the minimum distance from the test image. Each time one feature is used to select a training sample with the minimum distance and that sample is given one unit of score and this procedure should be performed for all features. In the end, the training sample with the highest score is selected as the most similar sample to the test subject.

Let us denote the $i$-th feature of the test sample by $\textbf{f}_i^{(t)}$, the predicted match for the test sample using this feature will be:
$$k^*(i)=\argmin_k \| \textbf{f}_i^{(t)}-\textbf{f}_i^{(k)} \|_2$$
where $\textbf{f}_i^{(k)}$ is $i$-th feature of the $k$-th person in the training data.

Let us denote the score of the $j$-th person based on $\textbf{f}_i$ by $S_j(i)$. $S_j(i)$ is equal to $\textrm{I}(j=\argmin_k\|\textbf{f}_i^{(t)}-\textbf{f}_i^{(k)}\|)$, where $\textrm{I}(x)$ denotes the indicator function. Then the total score of the $j$-th training sample using all the spectra is found by the following formula:
$$S_j=\sum_{All~spectra}\sum_{i=1}^{i_{max}}{\textrm{I}(j=\argmin_k|\textbf{f}_i^{(t)}-\textbf{f}_i^{(k)}|)}$$
Finishing the calculations, $j^*$ or the matched training sample will be:
\begin{gather*} 
j^*=\argmax_j \big[ S_j\big]= \argmax_j \big[\sum_{All~spectra}\sum_{i}S_j(i) \big]
\end{gather*}

\section{Results}
\label{SectionIV}

We have tested the proposed algorithm on the PolyU multisprectral palmprint database \cite{database} which has 6000 palmprints sampled from 500 persons (12 samples for each person). Each palmprint is taken under four different lights in two days resulting in a total of 24000 images. Each image is preprocessed and its ROI is extracted (with a size of 128 $\times$ 128). Images are acquired using four CCD cameras to take four images from each palmprint under four distinct lights: blue, green, red and near-infrared (NIR).

Before presenting the results, let us discuss briefly about the parameters of our model. 18 features are derived locally from blocks of size $16 \times 16$ (18 features for each block). Features of different blocks are concatenated resulting in a total of 1152 features for each image. For wavelet transform, Daubichies 2 is used. The recognition task is conducted using both majority voting and minimum distance classifier. Based on our experiment, majority voting algorithm achieves higher accuracy rate than minimum distance classifier and its result is used for comparison with other previous works.

We have studied the palmprint identification task for two different scenarios. In the first scenario, we have applied PCA to reduce the dimensionality of the feature space and used minimum distance classifier to perform template matching. The recognition accuracy for different number of PCA features is shown in Figure 3. As it can be seen, even by using 100 PCA features we are able to achieve very high accuracy rate.

\begin{figure}[3 h]

\begin{center}
    \includegraphics [scale=0.4] {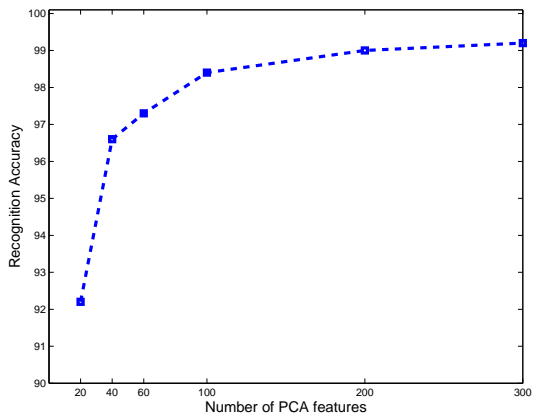}
\end{center}
\vspace{-0.4cm}
  \caption{Recognition accuracy for different number of PCA features}
\end{figure}

In the second scenario, we have used all 1152 features followed by weighted majority voting to perform palmprint recognition. Using all features enables us to achieve highly accurate results. The recognition rate for different fractions of training and testing data is shown in Table \ref{TblRes1}. For instance, in the case that the fraction of training sample is 4/12, we have used 2000 palmprints as training and the remaining 4000 ones as test samples. 

%The recognition task was also conducted using minimum distance classifier, but it showed a lower accuracy rate than majority voting algorithm and we did not report its result.
%The codes used for this purpose are online.

\begin{table} [h]
\centering
  \caption{Accuracy rate for different fraction of training samples }
  \centering
\begin{tabular}{|m{1.4cm}|m{1.5cm}|m{1.5cm}|m{1.5cm}|}
\hline
\centering{Training Fraction} & \centering{4/12} & \centering{5/12} & \ \ \ \ \ {6/12}\\
\hline
\ \ \centering{using all features} & \ \ ~~99.97\% & \ \ ~~100\% & \ \ ~~100\%\\
\hline
\end{tabular}
\label{TblRes1}
\end{table}

Table \ref{TblComp} provides a comparison of the results of our work and those of three other highly accurate schemes. The reported result for the proposed scheme corresponds to the case where all features are used and majority voting algorithm is employed for template matching. It can clearly be observed that the proposed method can perform better than the others which can be the result of the compatibility of the proposed features in this procedure.

\begin{table} [h]
\centering
  \caption{Comparison with other algorithms for palmprint recognition }
  \centering
\begin{tabular}{|m{6cm}|m{2cm}|}
\hline
\ \ \ \ \ \ \  Palmprint Recognition Schemes &  Recognition Rate\\
\hline
K-PCA+GWR \cite{PCA+GWR} & \ \ \ \ \ 95.17\% \\
\hline
Quaternion principal component analysis \cite{quaternion} & \ \ \ \ \ 98.13\% \\
\hline
Histogram of Oriented Lines \cite{hol} & \ \ \ \ \  99.97\% \\
\hline
Proposed scheme using majority voting algorithm & \ \ \ \ \  100\% \\
\hline
\end{tabular}
\label{TblComp}
\end{table}

%The experiment has been performed using MATLAB on a laptop with Windows 8 and Core i5 CPU running at 2.6GHz. The execution time for the proposed method is about 0.1s, 0.11s and 0.2s per test using DCT, wavelet and joint wavelet-DCT features respectively.

\section{Conclusion}
\label{SectionV}

This paper proposed a set of joint wavelet-DCT features for palmprint recognition. These features are extracted from non-overlapping sub-images so that they capture the local information of palmprints. These features are sensitive to the small differences between different palmprints. Therefore they are able to discriminate different palms with very similar patterns. After the features are extracted, PCA is applied for dimensionality reduction and majority voting algorithm is used to match each template to the most similar palmprint. 
The proposed algorithm has significant advantages over the previous popular approaches. Firstly, the proposed features here are very simple to extract and the algorithm is very fast to compute. Secondly, it has a very high accuracy rate for small fractions of training samples.
The same framework can be applied to other recognition tasks, such as fingerprint recognition and iris recognition.

\section*{Acknowledgments}
We would like to thank the Hong Kong Polytechnic University (PolyU) for sharing their multisprectral palmprint database with us.

%\section{REFERENCES}
\label{sec:ref}

%\bibliographystyle{IEEEbib}
%\bibliography{refs}
%\bibliography{ICIP_CVPR}

\end{document}